# A deep learning model for early prediction of Alzheimer's disease dementia based on hippocampal MRI

*Hongming Li[#], Mohamad Habes[#&], David A. Wolk[&], Yong Fan[#]*

*for the Alzheimer's Disease Neuroimaging Initiative and the Australian Imaging Biomarkers and Lifestyle Study of Aging* [*]

Section for Biomedical Image Analysis (SBIA), Center for Biomedical Image Computing and Analytics (CBICA), Department of Radiology[#] and Cognitive Neurology Division[&], Perelman School of Medicine, University of Pennsylvania, Philadelphia, PA, 19104, USA

Running title: Hippocampal MRI early predicts dementia

Correspondence to:

Yong Fan, Ph.D.

Richards Building, 7th floor, RM D703

3700 Hamilton Walk

Philadelphia, PA 19104-6116

Department of Radiology

Perelman School of Medicine

University of Pennsylvania

Philadelphia, PA 19104, USA

Email: yong.fan@uphs.upenn.edu; yong.fan@ieee.org

Tel: +1 215-746-4065

[*]Data used in preparation of this article were obtained from the Alzheimer's Disease Neuroimaging Initiative (ADNI) database (adni.loni.usc.edu) and the Australian Imaging Biomarkers and Lifestyle Study of Aging (AIBL) database (www.aibl.csiro.au). As such, the investigators within the ADNI and AIBL contributed to the design and implementation of ADNI and AIBL and/or provided data but did not participate in analysis or writing of this report. A complete listing of ADNI investigators can be found at: http://adni.loni.usc.edu/wp-content/uploads/how_to_apply/ADNI_Acknowledgement_List.pdf

# Abstract

**Introduction**: It is challenging at baseline to predict when and which individuals who meet criteria for mild cognitive impairment (MCI) will ultimately progress to Alzheimer's disease (AD) dementia.

**Methods**: A deep learning method is developed and validated based on MRI scans of 2146 subjects (803 for training and 1343 for validation) to predict MCI subjects' progression to AD dementia in a time-to-event analysis setting.

**Results**: The deep learning time-to-event model predicted individual subjects' progression to AD dementia with a concordance index (C-index) of 0.762 on 439 ADNI testing MCI subjects with follow-up duration from 6 to 78 months (quartiles: [24, 42, 54]) and a C-index of 0.781 on 40 AIBL testing MCI subjects with follow-up duration from 18-54 months (quartiles: [18, 36,54]). The predicted progression risk also clustered individual subjects into subgroups with significant differences in their progression time to AD dementia ($p<0.0002$). Improved performance for predicting progression to AD dementia (C-index=0.864) was obtained when the deep learning based progression risk was combined with baseline clinical measures.

**Conclusion:** Our method provides a cost effective and accurate means for prognosis and potentially to facilitate enrollment in clinical trials with individuals likely to progress within a specific temporal period.

**Background**

Individuals with mild cognitive impairment (MCI) are at a higher risk to develop dementia (usually due to Alzheimer's Disease (AD)), with an annual progression rate up to 10~20% [1]. Although clinical criteria for MCI and AD have been developed to formalize assessment of the gradual progression of the disease, it remains difficult at baseline to predict *when and which* individuals who meet criteria for MCI will ultimately progress to AD dementia.

The early prediction of AD dementia has been typically modeled as a pattern classification problem. For instance, by dichotomizing MCI subjects into progressive MCIs (pMCIs) and stable MCIs (sMCIs) based on a cut-off threshold of follow-up duration, a binary classifier is then trained based on baseline data to distinguish pMCIs from sMCIs. To early predict AD dementia based on neuroimaging data, machine learning techniques have been adopted to build classifiers upon imaging data, and prominent brain structural differences have been identified between AD and cognitively normal (NC) subjects as well as between pMCI and sMCI subjects within the medial temporal lobe (MTL), including regions such as hippocampus and entorhinal cortex [2-11]. The existing classification studies typically adopt relatively simple imaging measures, such as brain tissue density, volume, cortical thickness, and geometric characteristics of hippocampus. These hand-crafted measures might be less discriminative for the AD prognosis. For predicting MCI subjects' progression to AD dementia it is a suboptimal strategy to distinguish pMCI from sMCI subjects in a classification setting in that the classification performance is hindered on the cut-off threshold of follow-up duration to define pMCI and sMCI, and the cohorts of pMCI and sMCI subjects are highly heterogeneous regardless of the threshold used. More importantly, the classification based early prediction of AD dementia does not provide specific information about the timing of when MCI patients cross the threshold to AD dementia.

Recent studies have moved the focus onto the prediction of timing of progression to AD over the follow-up duration using time-to-event analysis techniques [12-16]. Clinical and imaging based measures at baseline and their longitudinal change trajectory have been adopted for predicting MCI subjects' progression to AD dementia, and promising performance have been achieved. However, only simple imaging features or individual clinical measures have been investigated, which might be less discriminative for the prognosis.

To better predict individual MCI subjects' progression to AD dementia based on baseline structural MRI data, we develop a deep learning framework to extract informative features from hippocampal MRI data, and build a prognostic model upon the extracted features to predict progression of MCI subjects in a time-to-event analysis setting. We have evaluated the proposed method using baseline structural MRI data of subjects from the Alzheimer's Disease Neuroimaging Initiative (ADNI) [17-19] (including ADNI-1, ADNI-GO&2), and the Australian Imaging Biomarkers and Lifestyle Study of Aging (AIBL) [20]. We also compared the deep learning based imaging features with conventionally hand-crafted imaging features.

**Methods**

*Imaging and clinical data*

We included data from the ADNI (http://adni.loni.usc.edu) and AIBL (www.aibl.csiro.au) cohorts, consisting of baseline MRI scans of 1711 ADNI subjects and 435 AIBL subjects. The data were downloaded on April 05, 2017. For up-to-date information, see www.adni-info.org.

We used MRI data (n=803 scans) from the ADNI-1 to train the proposed prognostic model. Then we validated the proposed model with independent data from the ADNI-GO&2 and the AIBL. For the ADNI-GO cohort only new add-on subjects (no overlap with ADNI-1) were used for the validation. The present study included all MCI subjects with baseline MRI scans and at least one clinical follow-up data point, including those who converted back from MCI to Normal Cognition. The characteristics of the cohorts included in this study are summarized in tables S1, S2, and S3 of supplementary material. In the present study, the ADNI-1 scans were collected using 1.5T scanners, the ADNI-GO&2 scans were collected using 3T scanners, and the AIBL scans were collected using 3T scanners.

Clinical variables include age, sex, education, APOE4 (Apolipoprotein E4), the 13-item version of the Alzheimer's Disease Assessment Scale-Cognitive subscale (ADAS-Cog13), Rey Auditory Verbal Learning Test (RAVLT) immediate, RAVLT learning, Functional Assessment Questionnaire (FAQ), and Mini-Mental State Examination (MMSE), were obtained for MCI subjects from the ADNI cohorts. Particularly, the ADAS-Cog13 consists of 11-item (word recall, commands, constructional praxis, naming, ideational praxis, orientation, word recognition, remembering test instructions, comprehension, word finding difficulty, spoken language ability) ADAS-cog plus 2 additional items (delayed word recall and number cancellation).

We also analyzed MCI subjects of the ADNI-GO&2 cohorts in terms of their amyloid positive status. Particularly, amyloid positive subjects were defined as those with a CSF (cerebrospinal fluid) Aβ42 (amyloid beta peptide 42) level below 192 pg/mL or with a summary AV-45 (Florbetapir-F18) cortical standardized uptake value ratio (SUVR) normalized by the whole cerebellum above 1.11 when CSF Aβ42 level was not available.

*Hippocampus extraction*

T1 MRI scans were registered to the MNI space using affine registration and resampled at a spatial resolution of $1 \times 1 \times 1$ mm$^3$. Bilateral hippocampal regions were segmented from the T1 images for each subject using the local label learning (LLL) [21] algorithm with 100 hippocampus atlases obtained from a preliminary release of the EADC-ADNI harmonized segmentation protocol project (www.hippocampal-protocol.net) [22]. A 3D bounding box of size $29 \times 21 \times 55$ was then adopted to extract hippocampus regions from the T1 images using the segmentation labels. These hippocampal data were used to extract features and build the prognostic models.

*Deep learning for informative feature extraction*

A deep learning model of convolutional neural networks (CNNs) with residual connections was trained to learning informative features for distinguishing AD from NC subjects. As illustrated in Fig. 1B, the left and right hippocampal images were inputs to the deep learning model with two streams, each for one hippocampus. Each stream of the deep learning model consisted of 1 convolutional layer (Conv), followed by 3 residual blocks (ResBlocks) [23], and 1 global average pooling (GAP) layer [24]. As illustrated in Fig. 1C, each of the ResBlocks had 2 Conv layers with a direct connection between its input and output. These two streams' outputs were then flattened and concatenated as input to a fully connected layer (FC) for building the classification model. Rectified linear units (ReLUs) were used as nonlinear activation functions and max pooling layers were adopted to learn features at multiple scales. Batch normalization (BN) was adopted in our deep learning model [25], and the GAP layers facilitated visualization of the learned features through AD-like relevance maps [24]. Specifically, each of the Conv layers contained 32 kernels, the ResBlocks 1, 2, and 3 contained 32, 64, and 128 kernels respectively, and all these kernels had the same size of $3 \times 3 \times 3$. The max pooling layer had a stride of $2 \times 2 \times 2$ and a kernel size of $2 \times 2 \times 2$. Outputs of the GAP layers were concatenated and followed by a dropout operation with a ratio of 0.5 to form an input vector for the FC to make the diagnosis of the input data. Once the deep classification model was obtained, new image could be fed into the deep learning model and its deep learning features ($1 \times 256$ feature vector in this study) were extracted as the output of the GAP layers and used as the input to time-to-event prognostic models.

*Time-to-event prognostic model based on deep imaging features*

Based on the deep learning features, a prognostic model for predicting individual subject's timing of progression to AD dementia was built using LASSO regularized Cox regression model [26]. Particularly, the prognostic model was trained based on the ADNI-1 cohort, and its prognostic performance was evaluated based on the ADNI-GO&2 and AIBL cohorts. The overall training and validation procedures are illustrated in Fig. 1A. The LASSO model's regularization parameter was optimized using 10-fold cross-validation based on the training data. The time-to-event prognostic model estimates overall risk scores to progress to AD dementia for individuals. Individuals with higher risk scores will progress to AD sooner than those with lower risk scores. An annual probability value of progression to AD can be estimated based on the risk score, given a baseline hazard (progression to AD) function, which could be estimated based on training cohort [27].

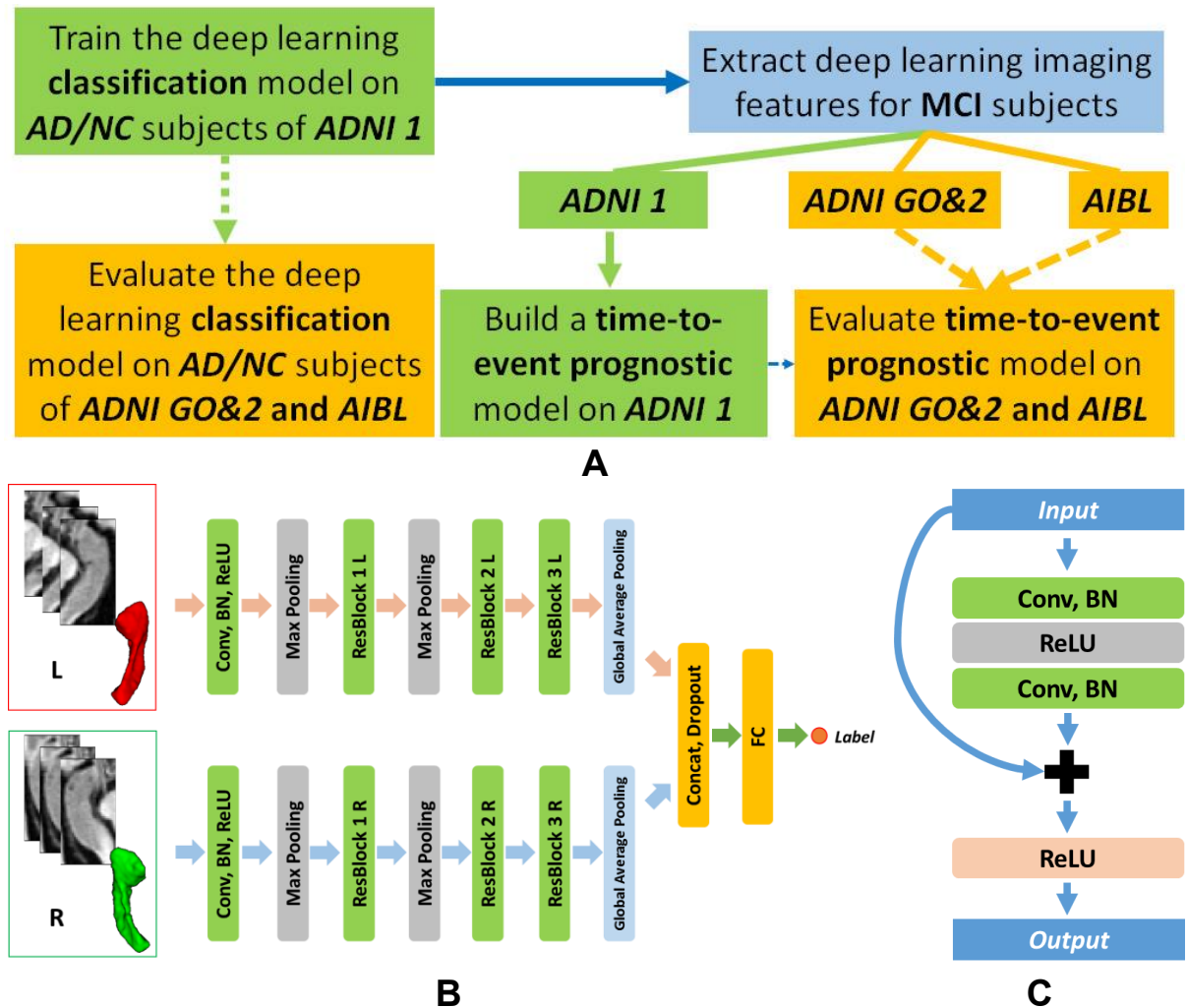

Fig. 1. Schematic illustration of the deep learning based AD prognosis. (A) A general flowchart for training and validating the prognostic model. (B) Deep network architecture for data-driven hippocampus-based AD diagnosis. (C) residual block. L: left hippocampus; R: right hippocampus. (NC: cognitively normal control; AD: Alzheimer's disease; MCI: mild cognitive impairment; Conv: convolutional layer; BN: batch normalization; ReLU: rectified linear unit; ResBlock: residual block; Concat: concatenation layer; FC: fully connected layer).

*Time-to-event prognostic model when deep imaging features meet clinical variables*

Based on the deep learning prognostic model, risk of progression to AD could be estimated for each MCI subject. The estimated risk was combined with clinical variables including age, sex, education, APOE4,

ADAS-Cog13, RAVLT immediate, RAVLT learning, FAQ, and MMSE to build a second prognostic model using Cox regression [28]. The prognostic model was trained based on the ADNI-1 MCI subjects and evaluated based on the ADNI-GO&2 data. Since the AIBL did not provide all these clinical measures, we did not evaluate the prognostic model based on the AIBL data.

To evaluate significance of the deep learning features when combined with the clinical measures for the prediction of AD dementia, we had also built another prognostic model of MCI subjects based on the AD score (probability) estimated by the deep AD/NC classification model and baseline clinical variables using Cox regression. The prognostic model was also trained on the ADNI-1 data and evaluated on the ADNI-GO&2 data.

**Validation and comparisons**

We evaluated the proposed method and compared it with state-of-the-art alternative image feature extraction methods [29] based on the same training and validation datasets. All the models were trained using the ADNI-1 data, and validated using the ADNI-GO&2 and AIBL data. We compared the deep learning models with those built on conventional hippocampal imaging features, including shape features and texture features (Details in supplementary material). Particularly, two alternative models were built on shape features only and a combination of shape and texture features (denoted as shape & texture thereafter) respectively.

In order to directly access reproducibility of our deep learning features across scanners with different magnetic field strengths, we obtained 1.5T and 3T scans of the same subjects of the ADNI-1 cohort (n=113, 37 NC, 50 MCI, and 26 AD), computed their deep learning features, and finally measured comparability between deep learning features of 1.5T and 3T scans and between their predicted risk scores for individual subjects.

In order to evaluate the deep imaging features' discriminative power, we compared its performance for distinguishing AD from NC subjects with alternative classification models based on conventional hippocampal shape and texture features. Random forests (RF) [30] were adopted to build classification models upon the conventional features (Details in supplementary material). Classification accuracy, receiver operating characteristic (ROC) curve, and area under ROC curve (AUC) were used to evaluate all the models under comparison. We adopted Delong test to compare AUC measures obtained by different models.

In order to investigate how different parts of the hippocampal imaging data contributed to the classification, class relevance maps [24] with respect to AD were obtained for all the subjects from the validation cohorts, and mean relevance maps of different subject groups including NC, sMCI, pMCI, and AD were obtained respectively. MCI subjects who progressed to AD from 0.5 to 3 years from the baseline scan were defined as pMCI (mean/standard deviation: 1.69/0.93 years, quartiles: [1, 2, 2.5] years), otherwise defined as sMCI for this visualization analysis. It is worth noting that we do not need to define pMCI and sMCI for the time-to-event analyses.

We further investigated the associations between the deep hippocampal imaging features and clinical measures including MMSE, ADAS-cog13, RAVLT immediate, RAVLT learning, FAQ, $A\beta$ status (positive or negative), and APOE4, Pearson's correlation coefficients were adopted to measure the potential associations.

In order to evaluate the performance of prognostic models built on different kinds of imaging features, concordance index (C-index) and time-dependent ROC curves were adopted to evaluate their accuracy. Particularly, the C-index measures proportion of all possible pairs of subjects, at least one of whom has progressed to AD dementia, in which the predicted progression risk (probability) is larger for the subjects who progressed to AD dementia in a shorter time [31], while the time-dependent ROC curves access prediction performance of progression of AD dementia at different observed times [32]. The time-dependent ROC curves were adopted so that the performance of prognostic models could be evaluated using ROC curves that are widely used to illustrate sensitivity and specificity of a continuous diagnostic marker for predicting a binary clinical outcome variable in classification studies. Since the status of progression of AD is dependent on the follow-up time, binary status of progression of AD could be obtained at different cut-off time points to compute their corresponding ROC curves, yielding time-dependent ROC curves [33]. A nonparametric approach [34] was adopted to estimate and compare C-index values, and nonparametric inverse probability of censoring weighting estimators [32] were adopted to obtain and compare time-dependent ROC based measures. The prognostic accuracy was also evaluated for amyloid positive MCI subjects alone. All the measures were calculated in R.

We investigated the subject stratification results based on the prognostic risks of progression to AD dementia for MCI subjects. Specifically, all the MCI subjects were categorized into 3 sub-groups with low, middle, and high predicted risks of progression to AD dementia, and Kaplan-Meier plot was adopted to investigate their progression to AD dementia based on real follow-up duration information.

**Results**

Mean of correlations between the deep learning features of 1.5T and 3T scans of the 113 ADNI-1 subjects was 0.955±0.023 (quantiles: [0.930, 0.958, 0.979]), and mean of intra-class correlation coefficients (ICCs) across all the deep learning features was 0.950±0.026 (quantiles: [0.925, 0.951, 0.978]), demonstrating that the deep learning features were robust to magnetic field strength differences. The ICC between the predicted risk scores of the 1.5T and 3T scans was 0.982, demonstrating that the prediction performance was also robust to the differences in magnetic field strength.

The deep learning classifier's accuracy rates for distinguishing AD from NC subjects on the ADNI-GO&2 and AIBL cohorts were 0.900 and 0.929 respectively, and AUC values were 0.956 and 0.958 respectively. Delong test indicated that the deep learning classifier performed significantly better than the RF based classifiers in terms of their AUC measures (p<0.008). Fig. S2 (supplementary material) shows ROC curves of the classifiers under comparison.

Fig. 2 illustrates the mean AD-like relevance maps for NC, sMCI, pMCI, and AD groups of the ADNI-GO&2 cohorts. We found that the relevance map of the AD subjects highlighted both the anterior and posterior hippocampus, the pMCI subjects' highlighted the anterior hippocampus, and maps of the NC and sMCI subjects had relatively weak relevance.

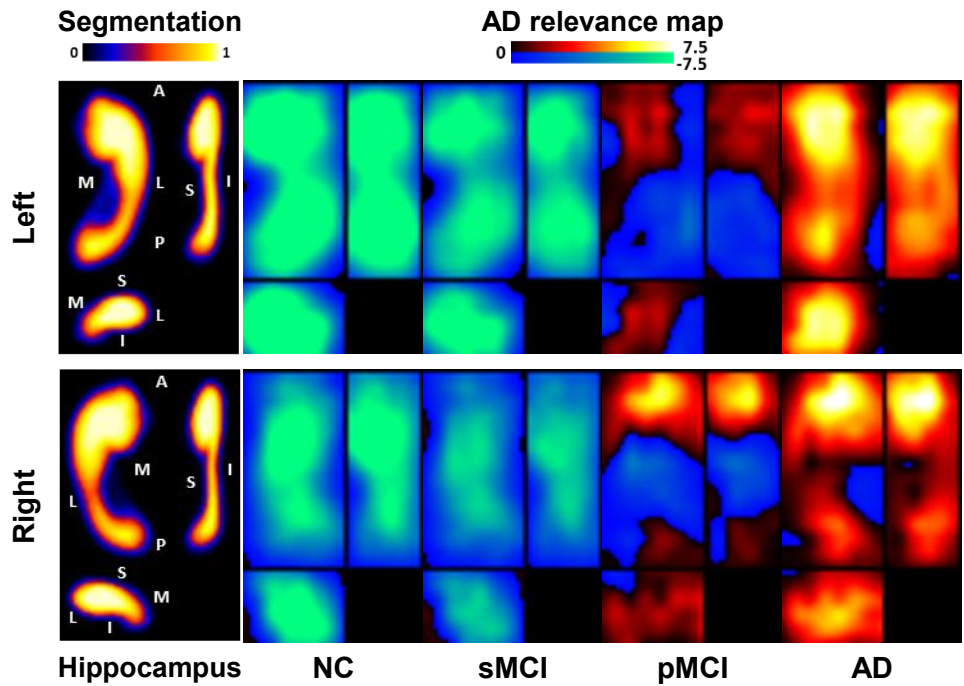

Fig. 2. Mean AD-like relevance maps for different subject groups, demonstrating the discriminative sub-regions that distinguish AD from NC and characterizing the imaging patterns along the pathological progression of AD. Warmer color indicates severer brain degeneration and more relevant to the progression to AD and cool color indicates rare brain degeneration and irrelevant to the progression to AD. The hippocampus panel shows mean hippocampal segmentation maps across subjects in different views (A: anterior, P: posterior, M: medial, L: lateral, S: superior, I: inferior). The mean maps have values in [0,1], indicating each voxel's percentage for being located in the hippocampus across subjects whose brain images were registered to the MNI space. The segmentation maps serve as spatial references to the hippocampal locations for the AD relevance maps. NC: cognitively normal control; AD: Alzheimer's disease; MCI: mild cognitive impairment; sMCI: stable MCI; pMCI: progressive MCI.

Fig. 3 shows that the correlation coefficients between top 50 deep learning features with largest weights in the deep learning classifier and cognitive measures and biomarkers across all subjects of the ADNI-GO&2 cohorts, indicating that the deep learning features were significantly correlated with clinical measures (Pearson's correlation, $p<0.05$). Fig. S3 (supplementary material) shows that amyloid negative MCI subjects had significantly lower progression risks compared with amyloid positive subjects ($p<6.0 \times$ $10^{-13}$, Wilcoxon rank sum tests). These results demonstrated that the deep imaging features were informative to capture the characteristics of AD related biomarkers and cognitive measures.

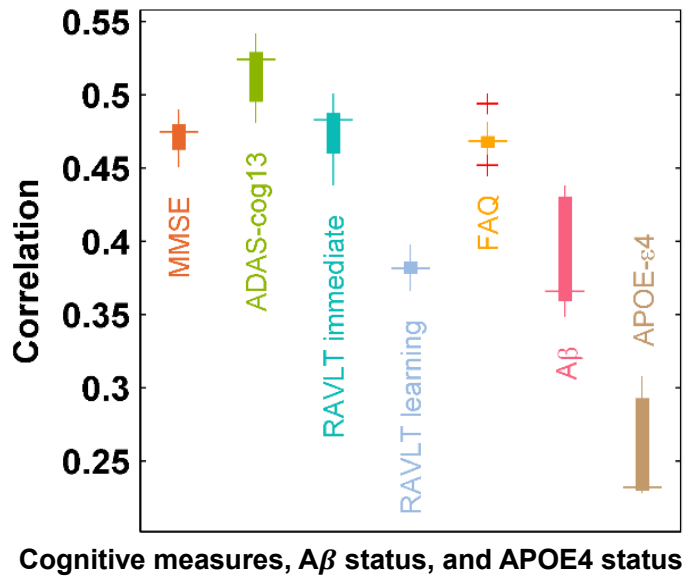

**Cognitive measures, A$\beta$ status, and APOE4 status**

Fig. 3. Top 50 features with largest weights in the deep classification model for AD/NC diagnosis were significantly correlated to cognitive/biological measures across all the subjects of the ADNI-GO&2 cohorts. The box plots show median/quartiles of correlation coefficients between pairs of one cognitive/biological measure and each of top 50 deep learning features. The outliers are plotted using the '+' symbol. (MMSE: Mini–Mental State Examination; ADAS-cog13: 13-item version of the Alzheimer's Disease Assessment Scale-Cognitive subscale; RAVLT: Rey Auditory Verbal Learning Test; FAQ: Functional Assessment Questionnaire; Aβ: amyloid beta peptide 42; APOE4: Apolipoprotein E4).

The prediction accuracy of the prognostic models built upon hippocampal imaging features are summarized in Table 1. The deep learning model predicted the ADNI-GO&2 MCI subjects' progression to AD dementia with a C-index of 0.762, significantly better than those built upon conventional shape and texture image features (p<0.03 [34]). The deep leaning model predicted the AIBL MCI subjects' progression to AD dementia with a C-index of 0.781, better than those built upon conventional shape (p=0.037). However, the difference between prediction models built on the deep learning imaging features and the shape & texture image features was not statistically significant (p=0.694 [34]). As shown in Fig.4 (top row), the AUC measures of time-dependent ROC curves obtained by the deep learning based model on follow-up durations from year 1 to year 3 were 0.75, 0.778, and 0.813 respectively on the ADNI-GO&2 cohorts, better than the alternative models (the differences were significant when compared with prediction models built on the shape features at year 2, and on both the shape and shape & texture features at year 3, p<0.04 [32]).

Table 1. Prediction performance of prognostic models built upon different types of features.

| C-index / Cohort | Hippocampal imaging features | | | Clinical measures | Clinical measures with deep learning imaging features |
| --- | --- | --- | --- | --- | --- |
| | Shape | Shape & Texture | Deep learning | | |
| **ADNI GO&2** | 0.676 (0.0006) | 0.707 (0.024) | **0.762** | 0.848 (0.05) | **0.864** |
| **AIBL** | 0.655 (0.037) | 0.762 (0.694) | **0.781** | | |

$p$ values between deep learning based model and alternatives are demonstrated in the parenthesis. The clinical measures included APOE4 status, all cognitive measures, and demographic data.

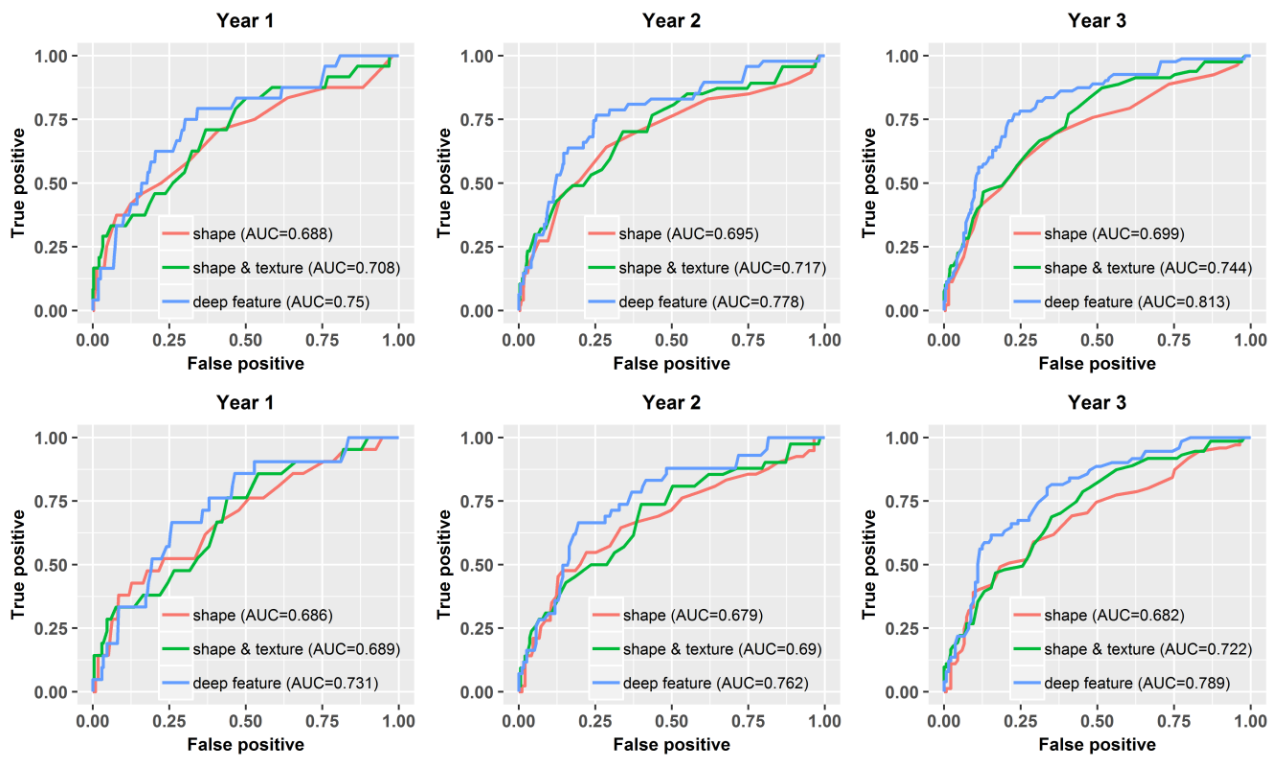

Fig. 4. Time-dependent ROC curves of prognostic models built upon different imaging features at follow-up durations from year 1 to year 3 on the ADNI-GO&2 cohorts. Top row: all the ADNI-GO&2 MCI subjects; Bottom row: Amyloid positive MCI subjects of the ADNI-GO&2 cohorts.

The prediction model built on the deep learning imaging features predicted the amyloid positive subjects' progression to AD with a C-index of 0.733, better than those built on the conventional shape and shape & texture imaging features (with C-index of 0.656 and 0.680 respectively, $p$=0.004 and 0.06 [34]). As shown in Fig.4 (bottom row), the AUC measures of time-dependent ROC curves obtained by the deep imaging feature based model at follow-up durations from year 1 to year 3 were 0.731, 0.762, and 0.789 respectively, better than the alternative models (the differences were significant when compared with prediction models built on the shape features at year 2 and 3, $p$<0.05 [32]). The deep learning feature based model also performed better than prediction models built upon hippocampal volumes (supplementary material).

The predicted progression risk to AD dementia also clustered MCI subjects from ADNI-GO&2 into subgroups with significant differences in their timing of progression (p<0.0002, Log-rank test) with age, sex, education and APOE4 status as covariates, as demonstrated by the Kaplan-Meier plots in Fig. 5.

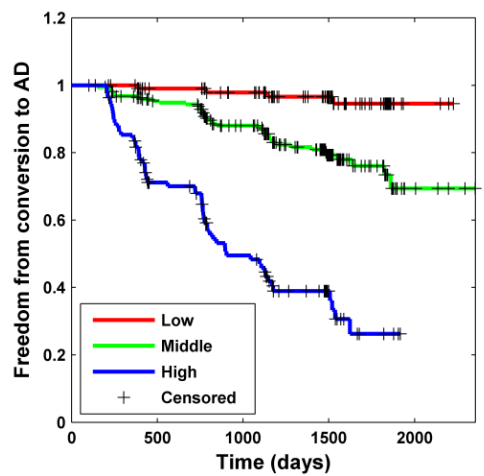

Fig. 5. Kaplan-Meier plots of MCI subgroups with different progression risks in terms of conversion to AD estimated by the deep learning prediction model on testing MCI subjects of the ADNI-GO&2 cohorts (Low: the 1st quartile, High: the 4th quartile, Middle: 2nd and 3rd quartiles). The subgroups were significantly different in their conversion timing to AD (Log-rank test, p<0.0002).

The prediction model built upon the predicted AD dementia progression risks (based on deep learning features) and baseline clinical measures achieved improved performance (C-index=0.864) on the ADNI-GO&2 cohorts, significantly better than the model built on clinical measures alone (C-index=0.848, p=0.05). As summarized in Table S4 and S5, RAVLT_immediate, FAQ, and the deep learning features were top 3 predictors for predicting progression to AD dementia (p<$1 \times 10^{-5}$). We have further estimated performance of prediction models built on different combinations of imaging features, Aβ measures, APOE4 status, and cognition measures, in addition to demographic data (age, sex, and education). As summarized in Tables S6 and S7 (supplementary material), the combination of the deep learning imaging features and demographic measures had significantly better prediction performance than the combination of MMSE and demographic measures (p=0.0004). Among the cognitive measures, ADAS-cog13 had the best prediction performance when combined with demographic measures, and the difference in terms of prediction performance between the deep learning imaging features and the ADAS-Cog13 was not statistically significant when combined with the demographic data (p=0.257). The deep learning imaging features had similar performance as A$\beta$ and APOE4 measures when combined with all the cognitive measures.

**Discussion**

In this study, we proposed a deep learning framework for early prognosis of AD dementia based on the hippocampal MRI data. We trained a deep learning classifier based on the ADNI-1 cohort for extracting informative imaging features, and built a time-to-event prognostic model on these features to predict the progression to AD dementia for MCI subjects of the ADNI-GO&2 and AIBL cohorts. We demonstrated that the deep learning prediction model could achieve promising performance for predicting MCI subjects' progression to AD dementia and identifying subgroups of subjects with different progression patterns.

Deep learning techniques have been explored for the prognosis of AD dementia [35-38]. These studies adopted a classification setting to predict MCI progression, and had to dichotomize the training data into progressive or stable MCIs based on certain cut-off threshold. Therefore, their prediction performance was dependent on their cut-off thresholds. Instead of formulating the early prediction of MCI subjects' progression to AD dementia as a binary classification problem, we built a prognostic model under a time-to-event analysis setting. The time-to-event prognostic model took into consideration the timing of progression to AD dementia, and could estimate the timing/risk of progression to AD dementia for each individual subject, which could be used to monitor their disease progression longitudinally. Our prediction results could also be analyzed using the conventional ROC curves, i.e., the time dependent ROC curves. On the other hand, the estimated risk could also facilitate stratification of MCI subjects to identify those with higher risk to progress to AD dementia as demonstrated in Fig. 5.

A very striking finding in our study was the clear advantage of the use of deep learning hippocampal features in predicting progression to dementia compared with the hippocampal volume, which is frequently used in the literatures as a marker of neurodegeneration [39]. Maybe due to practicality in assessment, most of the attention has been given so far for the volumetric features of the hippocampus rather than other features in exploring dementia. However, even in structural MRI the dementia related changes in the hippocampus seem to be better exploited with the deep learning features we are proposing for more robust progression prediction.

Several studies have specifically focused on the hippocampus for early diagnosis of AD and built predictive models upon anatomical features such as the hippocampal volume and shape based measures and image intensity texture features [5, 6, 40-44]. Although promising performance of the hippocampus shape features [45-47], texture features [48] and CNNs based features [43, 49] has been demonstrated for the classification of AD patients, their classification setting ignored the timing of progression to AD dementia. Moreover, it is challenging to define the pMCI and sMCI under a classification setting due to high heterogeneity of the AD continuum. Several recent studies [12-15] have focused on the prediction of time of progression to AD dementia under a time-to-event analysis setting; however, relatively simple features, e.g., volumetric and geometric measures, were included in the prognostic analysis. The discriminative power of these hand-crafted measures are relatively limited, especially when used for more complex prognostic tasks. As demonstrated in Fig. S2, the deep imaging features performed significantly better than conventional shape and texture features for distinguishing AD from NC subjects on both the ADNI-GO&2 and AIBL cohorts, indicating that the deep imaging features are more discriminative, and may have better potentials for characterizing the hippocampal changes related to AD dementia. It also demonstrated good generalization performance across different cohorts.

Sub-regions of the hippocampus, as shown in Fig. 2, contributed differently to characterize the AD related differences. The relevance map of the AD subjects highlighted both the anterior and posterior hippocampus, the pMCI subjects' map highlighted the anterior hippocampus, and maps of the NC and sMCI subjects had relatively weak relevance. These results were largely consistent with patterns of MCI and AD subjects described in existing studies [50, 51]. It also suggested that the anterior hippocampus was involved along the progression to AD prior to the posterior part [52, 53]. These relevance patterns have demonstrated that the deep features were indeed extracted from the AD related hippocampus regions. As the CNNs was optimized to learn discriminative imaging features, representing an evolution of imaging features from low-level intensity contrast to high-level complex patterns, for better differentiating AD from NC, it is speculated that the learned imaging features might reflect the AD relevant microstructural alternations in the hippocampal regions, and different weights of sub-regions demonstrated their involvement in these changes. These imaging alternations might be results of pathophysiological changes such as neuronal loss [54].

As shown in Fig. 3, we also found that the deep learning features were significantly correlated with cognitive measures and AD related biomarkers. Moreover, we found that amyloid positive MCIs who have molecular evidence of prodromal AD had higher predicted AD dementia progression risks than amyloid negative MCIs [55].

Different quantitative evaluation measures, as shown in Table 1 and Fig. 4, have demonstrated that the deep imaging features' superiority for predicting the MCI subjects' progression to AD on different cohorts, compared with the conventional shape and texture features. The deep learning prediction model also performed significantly better than other prediction models built upon conventional imaging features for amyloid positive subjects. Fig. 4 shows that the AUC values of the A$\beta$ positive MCI subjects were smaller than those of all MCI subjects. This difference in AUC values was a result of the difference in the testing data sets, and the risk scores of individual A$\beta$ positive MCI subjects were the same regardless of the computation of AUC values. Comparisons in terms of AUC values should be based on the same data set since the AUC values are summary measures for certain testing data sets. Nonetheless, it is possible that measures of neurodegeneration in the hippocampus would have better predictive power than amyloid status alone, as the former measure is more tightly linked to disease progression.

The prognostic performance of the prognostic model built on the combination of deep learning imaging features and clinical measures worked significantly better than that built on clinical measures alone. Particularly, RAVLT_immediate, FAQ, and the deep learning imaging features were top 3 predictors for predicting the MCI subjects' progression to AD dementia, as summarized in Table S4 and S5. The prognostic models built on a combination of demographic data (age, sex, education), cognitive measures, and deep learning features had better or similar prediction performance than prognostic models built on combinations of demographic data, cognitive measures, and APOE4 statuses or $A\beta$ measures as summarized in tables 1, S6, and S7, indicating that the deep learning imaging features could serve as a surrogate if APOE4 or $A\beta$ measures are not available.

Recently studies have also demonstrated promising performance for predicting individual subjects' timing of progression to AD dementia using time-to-event analysis techniques [12-15]. Particularly, clinical and imaging based measures at baseline [12, 13, 15] and their longitudinal change trajectory [14] have been adopted for predicting MCI subjects' progression to AD dementia. In conjunction with these studies, our results further demonstrated that the hippocampal MRI data could provide informative measures for the prediction of MCI subjects' timing of progression to AD dementia.

In the present study, the training imaging data were collected using 1.5T scanners and the testing imaging data were collected using 3T scanners. The deep learning features of 1.5 and 3T scans of the same ADNI 1 subjects highly correlated with each other and showed high comparability in their predicted risk scores, demonstrating that the deep learning features were robust to magnetic field strength differences. Therefore, the differences between the training data and testing data in the scanners' magnetic field strength minimally affected the prediction performance. We postulate that the image intensity normalization used in our study might minimize the difference caused by the scanner's magnetic field strength.

Our deep learning models could be used to predict AD progression risks for individuals with MRI scans collected following the ADNI imaging protocol. The whole pipeline of the proposed prognostic model is automatic. It is not sensitive to hippocampus segmentation, as only a bounding box containing hippocampus is required. The deep learning feature extraction and prognosis is efficient on both modern GPU and CPU (within 1 second) once the trained model is obtained. Particularly, for each subject, it takes about 0.011 second on a GPU or about 0.463 second on a CPU to compute deep learning features, and it takes 0.1 millisecond to obtain a prognosis result on a CPU. This deep learning tool can be used across platforms, including cloud computing, once they are containerized using docker.

The approach here offers a straightforward and computationally rapid means for a clinician to stratify MCI patients about the likelihood of progression within a particular timeframe. This could have significant impact on family and patient planning. In light of the fact that prediction accuracy is similar to prediction from Aβ measures, this approach may obviate the need for some measures in the clinical setting with the advantage of being non-invasive, as opposed to lumbar puncture, and less expensive, as compared to amyloid PET.

Although the proposed prognostic model has achieved better performance than state-of-the-art alternative imaging feature extraction methods for AD prognosis, further efforts are needed in following aspects. First, the current study focused on the hippocampus, it is expected to obtain improved prognostic performance when the deep learning method is applied to the whole brain MRI data. Second, the current study focused on predicting MCI subjects' progression to AD dementia, but the proposed framework could be applied to other clinical endpoints, such as predicting NC subject's progression to MCI [56] or cognitive decline, which could be useful in preclinical AD studies and facilitate subject screening in clinical trial. Third, only data at baseline were used in the current study, and we expect that the prognostic performance could be boosted if longitudinal data are incorporated into the model. Fourth, the current study focused on the prediction of boundaries between MCI and AD, which may not be well equipped to characterize the AD continuum [57].

The cognitive measures demonstrated better performance than the imaging measures for predicting the AD dementia progression as cognition is core component of the diagnosis of dementia, which results in circularity of using these measures in this type of prediction. Future work could focus more on predicting cognitive change which may be less susceptible to these circularity arguments. Additionally, the present study focused on the imaging features and did not fully explored other cognitive measures that might be even more informative for the prediction of MCI progression, such as RAVLT delayed recall, or impairment in other cognitive domains, such as executive function. Finally, changes in activities of daily living are also reflective progression from MCI to dementia and therefore could further add prediction.

Although our method could build time-to-event prediction models with continuous timing information, our prediction model largely captured the AD progression on intervals of 6 months because the training subjects were followed every 6 months. Since most of the ADNI MCI subjects progressed to AD dementia within 36 months, our prediction model might be driven to focus more on the advanced MCI subjects and to identify information relevant to late MCI. Moreover, although quantitative results have demonstrated that the prognostic model was robust to imaging data collected by scanners with different magnetic field strength, other potential confounders regarding subject heterogeneity such as atypical forms of AD require further investigation using datasets in clinical settings. One source of comfort related to heterogeneity of imaging acquisition for the viability of this approach is that quantitative results have demonstrated that the deep learning imaging features were robust to imaging data collected using scanners with different magnetic field strengths. It should be noted also that ADNI was designed to mimic a clinical trial and that is an additional context, outside of clinical practice, in which the findings here are relevant for potentially determining inclusion in an intervention study. Finally, although our study showed that the deep hippocampal features have considerable advantages compared to conventional methods such as hippocampal volume, the hippocampus as a structure might be limited in sensitivity and specificity for AD prediction and future research should consider further relevant regions.

In conclusion, the deep learning method for early prognosis of AD dementia could achieve promising performance, help distinguish MCI subjects with different progression patterns, and identify MCI subjects with higher risk to develop AD dementia, thus providing a cost effective and accurate means for prognosis and potentially to facilitate enrollment in clinical trials with individuals likely to progress within a specific temporal period.

## Acknowledgement

This work was supported in part by National Institutes of Health grants (Nos. EB022573, CA189523, DA039215, AG054409, and HL127659-04S1).

Data collection and sharing for this project was funded by the Alzheimer's Disease Neuroimaging Initiative (ADNI) (National Institutes of Health Grant U01 AG024904) and DOD ADNI (Department of Defense award number W81XWH-12-2-0012). ADNI is funded by the National Institute on Aging, the National Institute of Biomedical Imaging and Bioengineering, and through generous contributions from the following: AbbVie, Alzheimer's Association; Alzheimer's Drug Discovery Foundation; Araclon Biotech; BioClinica, Inc.; Biogen; Bristol-Myers Squibb Company; CereSpir, Inc.; Cogstate; Eisai Inc.; Elan Pharmaceuticals, Inc.; Eli Lilly and Company; EuroImmun; F. Hoffmann-La Roche Ltd and its affiliated company Genentech, Inc.; Fujirebio; GE Healthcare; IXICO Ltd.; Janssen Alzheimer Immunotherapy Research & Development, LLC.; Johnson & Johnson Pharmaceutical Research & Development LLC.; Lumosity; Lundbeck; Merck & Co., Inc.; Meso Scale Diagnostics, LLC.; NeuroRx Research; Neurotrack Technologies; Novartis Pharmaceuticals Corporation; Pfizer Inc.; Piramal Imaging; Servier; Takeda Pharmaceutical Company; and Transition Therapeutics. The Canadian Institutes of Health Research is providing funds to support ADNI clinical sites in Canada. Private sector contributions are facilitated by the Foundation for the National Institutes of Health (www.fnih.org). The grantee organization is the Northern California Institute for Research and Education, and the study is coordinated by the Alzheimer's

Therapeutic Research Institute at the University of Southern California. ADNI data are disseminated by the Laboratory for Neuro Imaging at the University of Southern California.

## Competing interests

Dr. Wolk reported receiving grants and personal fees for consultation from GE Healthcare, Merck, Eli Lilly, and Jannsen during the conduct of the study; other authors report no competing interests.

# A deep learning model for early prediction of Alzheimer's disease dementia based on hippocampal MRI

# (supplementary material)

*Hongming Li[#], Mohamad Habes[#&], David A. Wolk[&], Yong Fan[#]*
*for the Alzheimer's Disease Neuroimaging Initiative and the Australian Imaging Biomarkers and Lifestyle Study of Aging*[*]

Section for Biomedical Image Analysis (SBIA), Center for Biomedical Image Computing and Analytics (CBICA), Department of Radiology[#] and Cognitive Neurology Division[&], Perelman School of Medicine, University of Pennsylvania, Philadelphia, PA, 19104, USA

## Data

In the present study, the ADNI-1 scans were collected using 1.5T scanners, the ADNI-GO&2 scans were collected using 3T scanners, and the AIBL scans were collected using 3T scanners. The AD diagnosis was based on the National Institute of Neurological and Communicative Disorders and Stroke (NINCDS) and the Alzheimer's Disease and Related Disorders Association (ADRDA) clinical criteria for probable AD. Demographic information of the AD and NC subjects is summarized in Table S1. Compared with the ADNI-1 cohort, both the ADNI-GO&2 and the AIBL NC subjects were younger, and the AIBL NC and AD subjects had smaller MMSE.

Table S1. Demographic information of the NC and AD subjects included in this study.

| NC and AD subjects | | | |
|---|---|---|---|
| **Cohorts** | | **NC** | **AD** |
| **ADNI-1** | Number of subjects | 228 | 192 |
| | Gender (M/F) | 118/110 | 101/91 |
| | Age | 75.97±5.02 | 75.34±7.45 |
| | MMSE | 29.11±1.00 | 23.31±2.04 |
| **ADNI-GO&2** | Number of subjects | 311 | 158 |
| | Gender (M/F) | 142/169 | 91/67 |
| | Age | 72.98±6.09* | 74.85±8.09 |
| | MMSE | 29.00±1.25 | 23.1±2.07 |
| **AIBL** | Number of subjects | 328 | 67 |
| | Gender (M/F) | 145/183 | 26/41 |
| | Age | 72.30±6.51* | 73.28±8.08 |
| | MMSE | 28.71±1.21* | 20.36±5.67* |

NC: cognitively normal control; AD: Alzheimer's disease

*: $p < 0.05$, Wilcoxon rank sum test for Age and MMSE, Chi-Squared test for Gender.

---

*Data used in preparation of this article were obtained from the Alzheimer's Disease Neuroimaging Initiative (ADNI) database (adni.loni.usc.edu) and the Australian Imaging Biomarkers and Lifestyle Study of Aging (AIBL) database (www.aibl.csiro.au). As such, the investigators within the ADNI and AIBL contributed to the design and implementation of ADNI and AIBL and/or provided data but did not participate in analysis or writing of this report. A complete listing of ADNI investigators can be found at: http://adni.loni.usc.edu/wp-content/uploads/how_to_apply/ADNI_Acknowledgement_List.pdf

Most of the ADNI MCI subjects were followed every 6 months, and the AIBL MCI subjects were followed every 18 months. The follow-up information of the MCI subjects is summarized in Table S2. Compared with the ADNI-1 MCI subjects, the ADNI-GO&2 MCI subjects were younger with a different gender ratio and had larger MMSE. No significant difference was observed between MCI subjects of the ADNI-1 and the AIBL cohorts.

Table S2. Demographic and follow-up information of the MCI subjects included in this study.

| | MCI subjects with follow-up | | | | | | |
|---|---|---|---|---|---|---|---|
| Cohorts | Number of subjects | Gender (M/F) | Age (years) | MMSE | Follow-up months (quartiles) | Progressors | Months to progression (quartiles) |
| ADNI-1 | 383 | 246/137 | 74.73±7.32 | 27.01±1.77 | 6-132 ([12, 24, 48]) | 208 | 6-126 ([12, 24, 36]) |
| ADNI-GO&2 | 439 | 242/197* | 71.71±7.36* | 28.06±1.70* | 6-78 ([24, 42, 54]) | 107 | 6-60 ([12, 24, 36]) |
| AIBL | 40 | 21/19 | 76.13±7.28 | 27.25±2.10 | 18-54 ([18, 36, 54]) | 18 | 18-54 ([18, 18, 36]) |

MCI: mild cognitive impairment; Converter: MCI subjects who progressed to AD dementia before the last follow-up visit.
*: $p < 0.05$, Wilcoxon rank sum test for Age and MMSE, Chi-Squared test for Gender.

*Training of the deep classification model (AD versus NC) for imaging feature extraction*

The deep classification model was optimized using stochastic gradient descent algorithm [1], the momentum was set to 0.9, and the base learning rate was set to 0.01. The learning rate was updated using a stepwise policy, which drops the learning rate by a factor of 0.1 after every 20000 steps. The maximum iteration of the training procedure was set to 100000. Batch size of 32 was adopted to update weights in the model. The deep learning models were implemented based on Caffe [2], and trained on a Nvidia Titan X (Pascal) graphics processing unit (GPU). The classification model was trained using the ADNI 1 cohort, and validated using the ADNI-GO&2 cohorts.

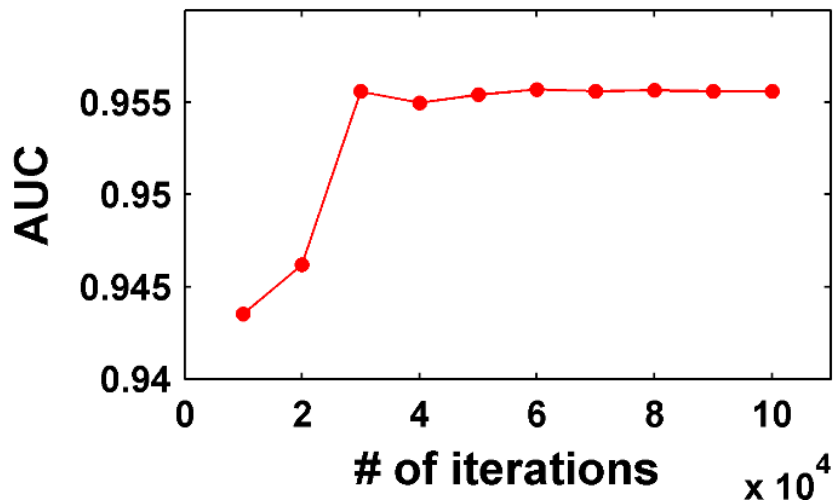

Fig. S1. Classification performance on the evaluation dataset with different number of train iteration steps.

To boost the deep learning model's performance and robustness, data augmentation is adopted to generate more training data [3]. Particularly, augmented image data were generated using image translation and non-rigid deformable image registration techniques. In particular, each hippocampus image along with its corresponding hippocampus masks in the training dataset was translated by 2 voxels along 26 directions of 3D image space separately, yielding augmented images that account for translation invariance for training the deep learning model. A non-rigid deformable image registration method, namely ANTs [4], was adopted with its default parameter setting to register one hippocampal MRI image, referred to as moving image, to another of the same side (left to left and right to right) within the same disease category (NC to NC, and AD to AD), and the resulting deformation field was used to deform the moving hippocampus image and its hippocampus label to generate deformed hippocampus image and label. In total, 10920 spatial translated images, and 44214 non-rigid registered images were generated for training the deep learning model.

To identify the optimal number of training iteration steps, classification performance (area under the ROC curve, AUC) of the trained models with different numbers of training iteration steps on the evaluation dataset (the ADNI-GO&2) were investigated as shown in Fig. S1, the classification model with the highest AUC (with 60000 training iterations) was adopted for imaging feature extraction in this study.

Fig. S2 shows ROC curves of the binary classifiers for distinguishing AD patients from NC subjects, obtained based on the evaluation cohorts using the shape, shape & texture, and deep imaging features respectively. The classification accuracy rate obtained by the deep imaging based binary classifier was 0.900 and 0.929 on the ADNI-GO&2 and AIBL cohorts respectively, while those obtained by the RF classifiers built upon hippocampal shape features and a combination of shape and texture features were 0.829 and 0.866 respectively on the ADNI-GO&2 cohorts, and 0.801 and 0.876 respectively on the AIBL cohort. The AUC values obtained by the deep imaging classifier were 0.956 and 0.958 on the ADNI-GO&2 and AIBL cohorts respectively, while those obtained by the RF classifiers built upon the hippocampal shape features and a combination of shape and texture features were 0.886 and 0.913 respectively on the ADNI-GO&2 cohorts, and 0.853 and 0.883 respectively on the AIBL cohort. Delong test indicated that the deep imaging classifier performed significantly better than the RF classifiers in terms of their AUC measures ($p<0.008$).

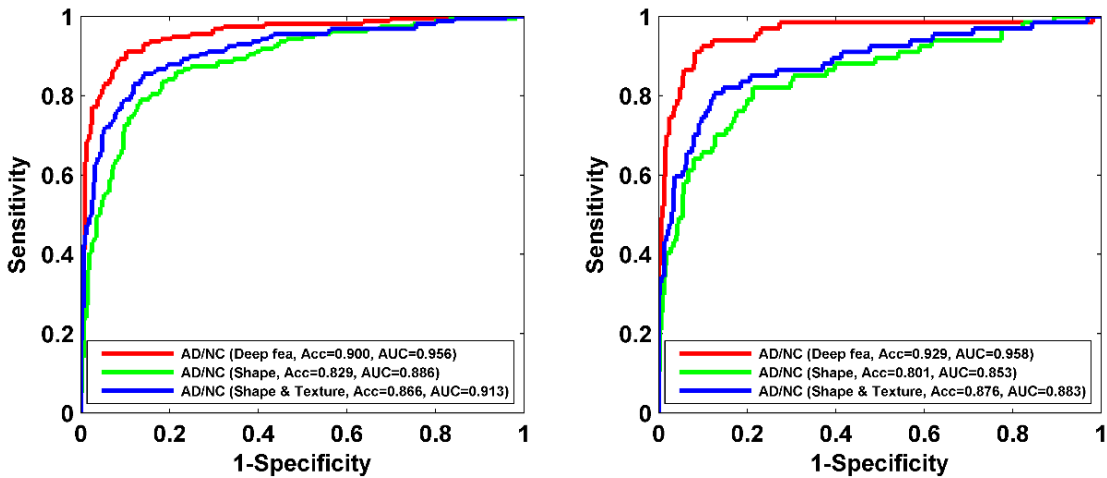

Fig. S2. ROC curves and classification accuracy (Acc) rates obtained by different binary classifiers for distinguishing AD patients from NC subjects, including the CNNs based classifier, and RF classifiers built upon the hippocampal shape and texture features, on the ADNI-GO&2 (left) and the AIBL cohorts (right).

*Comparisons with state-of-the-art imaging features*

We compared the deep learning based imaging features with state-of-the-art feature extraction methods for hippocampal MRI images. The feature extraction methods under comparison include conventional hippocampal shape and texture features. The conventional hippocampal shape and texture features were extracted using a software package (http://pyradiomics.readthedocs.io) [5]. Particularly, the hippocampal shape features were extracted from each subject's left and right hippocampal segmentation labels separately, including volume, maximum 3D diameter, maximum 2D diameter (column, row, and slice), surface area, surface volume ratio, flatness, sphericity, elongation, and spherical disproportion. In total, 22 shape features were extracted for each subject. The hippocampal texture features were extracted from the hippocampal images and their counterparts after wavelet decomposition, including the first order features, gray level co-occurrence matrix features, gray level size zone matrix features, and gray level run length matrix features. In total, 711 textures were extracted from each hippocampus region and 1422 hippocampal texture features were extracted from each subject.

To build the RF classifiers for AD/NC diagnosis, 1000 decision trees were adopted, and the minimum leaf size of the tree was set to 5. Sample weight for each training image was set to the ratio between total number of training images and the number of images within the same category, and the training images were sampled with replacement during the training procedure. The built-in RF implementation, namely TreeBagger, in Matlab (R2013a) was adopted to train the RF classifiers, and default values were used for other parameters.

### *Predicted risks of progression to AD dementia for MCI subjects with different amyloid positive statuses*

In addition to prognostic accuracy, we compared the predicted AD progression risks between amyloid positive and negative MCI subjects of the ADNI-GO&2 cohorts. As illustrated in Fig. S3, MCI subjects with A$\beta$-negative status had significantly lower progression risks compared with A$\beta$-positive subjects, no matter whether A$\beta$ status was defined based on CSF measure only, PET measure only, or their combination ($p$=$3.21 \times 10^{-13}$, $5.33 \times 10^{-13}$, $4.14 \times 10^{-14}$ respectively, Wilcoxon rank sum test).

Table S3. Amyloid positive and negative MCI subjects of the ADNI-GO&2 cohorts.

| | MCI subjects (n=439) | | |
|---|---|---|---|
| | CSF A$\beta$ (positive) | SUVR-1.11 (positive) | combination (positive) |
| Available subjects | 410 (255) | 436 (242) | 439 (271) |

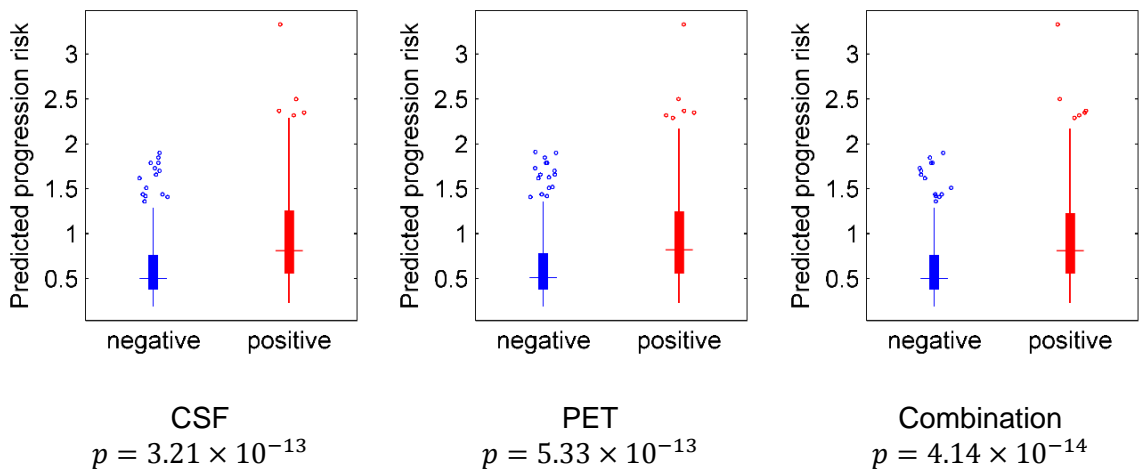

CSF  
$p = 3.21 \times 10^{-13}$

PET  
$p = 5.33 \times 10^{-13}$

Combination  
$p = 4.14 \times 10^{-14}$

Fig. S3. Predicted risks of progression to AD dementia of the ADNI-GO&2 MCI subjects with different amyloid positive statuses. The amyloid status was defined based on CSF measure only, PET measure only, and their combination.

### *When imaging features meet clinical variables for prognosis*

Beyond the AD progression risks estimated based on the deep learning imaging features, we have also built a second prognostic model of MCI subjects based on a combination of AD classification scores (probabilities) estimated by the deep learning AD/NC classification model and baseline clinical measures as described previously using Cox regression [6]. The prognostic model was also trained on the ADNI-1 cohort and evaluated on the ADNI-GO&2 cohorts. Details of the obtained Cox regression model are demonstrated in Table S4. The model obtained similar performance as that obtained by the AD progression risk based prognostic model, with a C-index 0.861. The AD classification score was significantly associated with the prognosis as demonstrated in Table S4, indicating that it could provide complimentary information when combined with clinical measures for the prognosis.

Table S4. Cox regression model of progression to AD dementia of MCI subjects, built on a combination of AD classification score estimated by the deep learning AD/NC classification model and baseline clinical measures.

| Variables | coef | Hazard ratio: *Exp*(coef) | Se(coef) | z | p |
|---|---|---|---|---|---|
| Age | -0.071 | 0.931 | 0.077 | -0.93 | 0.354 |
| Sex | 0.144 | 1.155 | 0.078 | 1.86 | 0.063 |
| Education | 0.109 | 1.116 | 0.073 | 1.49 | 0.135 |
| APOE4 | 0.166 | 1.180 | 0.075 | 2.22 | 0.026 |
| ADAS-Cog13 | 0.206 | 1.229 | 0.104 | 1.98 | 0.048 |
| RAVLT_immediate | -0.495 | 0.610 | 0.109 | -4.54 | 5.6e-06 |
| RAVLT_learning | 0.095 | 1.100 | 0.092 | 1.03 | 0.301 |
| FAQ | 0.340 | 1.404 | 0.060 | 5.69 | 1.3e-08 |
| MMSE | -0.102 | 0.903 | 0.080 | -1.28 | 0.200 |
| Deep learning imaging feature based AD score | 0.405 | 1.499 | 0.090 | 4.51 | 6.4e-06 |
| Likelihood ratio test=177.6 on 10 df, p<2e-16, n=383, number of events=208 | | | | | |

ADAS-Cog13: the 13-item version of the Alzheimer's Disease Assessment Scale-Cognitive subscale; RAVLT: Rey Auditory Verbal Learning Test; FAQ: Functional Assessment Questionnaire; MMSE: Mini-Mental State Examination.

Table S5 summarizes the Cox regression model of progression to AD of MCI subjects built on baseline clinical variables and the prognostic risk of progression to AD dementia derived from the deep learning imaging features. Since the prognostic risk of progression to AD dementia derived from the deep learning imaging features were derived based on the ADNI-1 MCI subjects, its p-value was subject to a circular analysis risk. Therefore, we evaluated the deep leaning based imaging features based on the AD classification scores which were derived from the AD and NC subjects, independent on the MCI subjects, as summarized in Table S4.

Table S5. Cox regression model of conversion to AD of MCI subjects when combining deep imaging feature based risk of conversion to AD dementia and baseline clinical variables.

| Variables | coef | Hazard ratio: *Exp*(coef) | Se(coef) | z | p |
|---|---|---|---|---|---|
| Age | -0.1615 | 0.8509 | 0.0803 | -2.01 | 0.044 |
| Sex | 0.1266 | 1.1350 | 0.0771 | 1.64 | 0.101 |
| Education | 0.0668 | 1.0691 | 0.0740 | 0.90 | 0.367 |
| APOE4 | 0.1334 | 1.1427 | 0.0743 | 1.80 | 0.073 |
| ADAS-Cog13 | 0.1567 | 1.1696 | 0.1040 | 1.51 | 0.132 |
| RAVLT_immediate | -0.5067 | 0.6025 | 0.1098 | -4.61 | 3.9e-06 |
| RAVLT_learning | 0.0895 | 1.0937 | 0.0907 | 0.99 | 0.324 |
| FAQ | 0.3358 | 1.3990 | 0.0610 | 5.51 | 3.6e-08 |
| MMSE | -0.0737 | 0.9290 | 0.0812 | -0.91 | 0.364 |
| Deep learning imaging feature based the prognostic risk | 0.5430 | 1.7211 | 0.0894 | 6.07 | 1.3e-09 |
| Likelihood ratio test=192.1 on 10 df, p=<2e-16, n= 383, number of events= 208 | | | | | |

As shown in Fig. S4, the prediction model built on a combination of the deep learning imaging features and clinical measures obtained time-dependent AUC measures of 0.885, 0.892, and 0.916 at follow-up durations from year 1 to year 3 respectively, better than the prediction model built upon the clinical measures alone (p=0.466, 0.054, and 0.067 at years 1 to 3 respectively).

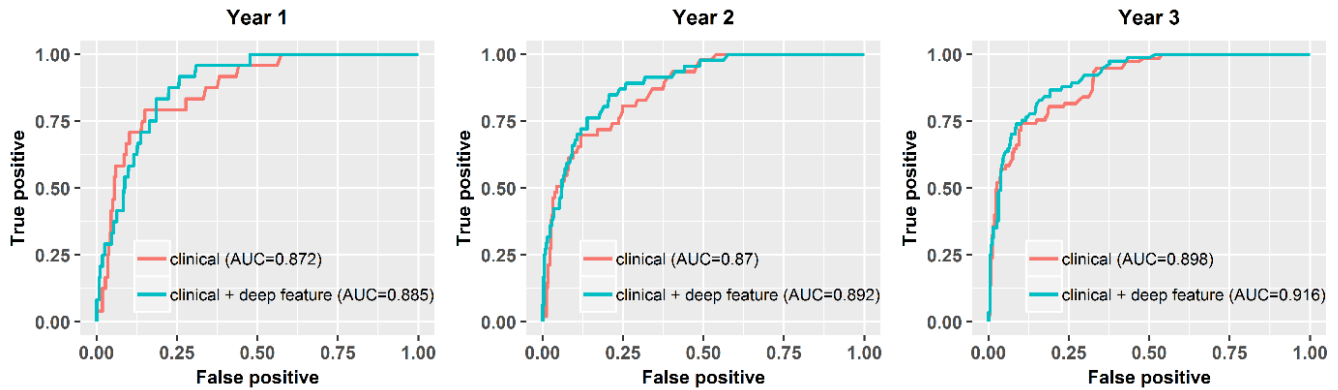

Fig. S4. Time-dependent ROC curves of prognostic models built upon baseline clinical measures and combination of clinical measures and the prognostic risk of conversion to AD derived from deep learning features at follow-up durations from year 1 to year 3 on the ADNI-GO&2 cohorts.

We have further evaluated prediction models built on different combinations of imaging features, CSF A$\beta$ measures, APOE4 status, cognitive measures, and demographic data (age, sex, and education), in addition to the evaluation of prediction models built on imaging features, clinical measures (APOE4 status, all cognitive measures, and demographic data), and their combination. All the prediction models were built on the ADNI-1 cohort and evaluated based on the ANDI-GO&2 cohorts and the AIBL cohort. We have not evaluated the prediction models built on the clinical measures or CSF A$\beta$ measures on the AIBL data set because the AIBL cohort and the ADNI cohorts had different clinical measures and we do not have access to the AIBL CSF A$\beta$ measures.

For A$\beta$ measures, we focused on CSF measures for its wide availability in the ADNI 1, 2, and GO cohorts. Since CSF A$\beta$ measures were not available for a proportion of the MCI subjects (6.6%), we had two different pairs of training/testing data sets. One pair of the training/testing data sets consist of subjects with imaging features and clinical measures including APOE4 status, cognitive measures, and demographic data (CSF A$\beta$ measure was not considered). This pair of data sets were used to obtain results summarized in Table 1 and Table S6. The other pair of the training/testing data sets consist of subjects with all imaging features, clinical measures aforementioned, and CSF A$\beta$ measure (subjects without CSF A$\beta$ measures were excluded). Prediction performance measures obtained based on this pair of data sets are summarized in Table S7.

Table S6 summarizes performance measures of prediction models built on imaging features, APOE4 status, cognitive measures, demographic data, and their combinations (without taking into consideration A$\beta$ measures). The demographic measures were included in all the prediction models. Each element of Table S6 shows C-index values obtained by a prediction model built on measures shown in its corresponding row and column. Particularly, the combination of the deep learning imaging features and demographic measures had significantly better prediction performance than the combination of MMSE and demographic measures (p=0.0004). Among the cognitive measures, ADAS-cog 13 had the best prediction performance when combined with demographic measures, and the difference in terms of prediction performance between the deep learning imaging features and the ADAS-Cog13 was not statistically significant when combined with the demographic data (p=0.257).

The combination of cognitive measures had the overall best prediction performance. This result is not surprising as cognition is core component of the diagnosis of dementia, and therefore the prediction performance of cognitive measures for progression to dementia likely reflect some circularity.

Table S6. Prediction performance of models built on different combinations of demographic data, cognitive measures, APOE4, and different types of imaging features.

| C-index | Demographic | +ADAS13 | +RAVLT immediate | +RAVLT learning | +FAQ | +MMSE | +all cognitive measures |
|---|---|---|---|---|---|---|---|
| Demographic | 0.478 | 0.789 | 0.754 | 0.665 | 0.770 | 0.644 | 0.842 |
| +APOE4 | 0.652 | 0.801 | 0.778 | 0.710 | 0.778 | 0.703 | 0.848 |
| +Shape | 0.665 | 0.803 | 0.783 | 0.715 | 0.779 | 0.707 | 0.847 |
| +Shape & Texture | 0.699 | 0.808 | 0.797 | 0.729 | 0.789 | 0.736 | 0.852 |
| +Deep Image risk (DIR) | 0.763 | 0.818 | 0.825 | 0.772 | 0.811 | 0.779 | 0.858 |
| +DIR & APOE4 | 0.777 | 0.825 | 0.832 | 0.786 | 0.824 | 0.792 | 0.864 |

Table S7 summarizes performance measures of the prediction models built on imaging features, CSF A$\beta$ measures, APOE4 status, cognitive measures, demographic measures, and their combinations. Both A$\beta$ status (positive or negative) and A$\beta$ value were evaluated. Similarly, the demographic measures were included in all the prediction models, and each element of Table S7 shows C-index values obtained by a prediction model built on measures shown in its corresponding row and column. Particularly, the deep learning imaging features had better performance than the CSF A$\beta$ measures (status or value) and APOE4 status (the first column of Table S7) when combined with the demographic data. However, they had similar prediction performance if they were combined with all cognitive measures (the last column of Table S7), indicating that the deep learning imaging features might serve as a surrogate for these measures. Furthermore, prediction models built on individual cognitive measures did not obtain significantly better prediction performance than the prediction model built on the deep learning imaging features when combined with the demographic data.

Table S7. Prediction performance of models built on different combinations of demographic data, cognitive measures, APOE4, CSF A$\beta$ measures (status: positive/negative, or continuous measure) and different types of imaging features. (p values between prognostic models based on demographic data & deep image risk and demographic data & single alternative clinical measure or combination of APOE4 and A$\beta$ measure are demonstrated in the parenthesis).

| C-index | Demographic | +ADAS13 | +RAVLT immediate | +RAVLT learning | +FAQ | +MMSE | +all cognitive measures |
|---|---|---|---|---|---|---|---|
| Demographic | 0.516 | 0.779 (0.425) | 0.741 (0.518) | 0.663 (0.0009) | 0.762 (0.962) | 0.649 (0.0006) | 0.819 |
| +APOE4 | 0.644 (0.0002) | 0.791 | 0.761 | 0.695 | 0.779 | 0.691 | 0.827 |
| +A$\beta$ status | 0.668 (0.008) | 0.804 | 0.780 | 0.735 | 0.793 | 0.734 | 0.836 |
| + A$\beta$ value | 0.701 (0.03) | 0.801 | 0.775 | 0.727 | 0.789 | 0.735 | 0.831 |
| + A$\beta$ status & APOE4 | 0.683 (0.004) | 0.805 | 0.783 | 0.738 | 0.798 | 0.737 | 0.838 |
| + A$\beta$ value & APOE4 | 0.706 (0.045) | 0.802 | 0.778 | 0.727 | 0.791 | 0.736 | 0.833 |
| +Shape | 0.673 (0.0008) | 0.795 | 0.777 | 0.717 | 0.780 | 0.712 | 0.824 |
| +Shape & Texture | 0.690 (0.002) | 0.796 | 0.781 | 0.725 | 0.779 | 0.729 | 0.828 |
| +Deep Image risk (DIR) | 0.759 | 0.805 | 0.809 | 0.767 | 0.809 | 0.773 | 0.840 |
| +DIR & A$\beta$ status | 0.796 | 0.822 | 0.823 | 0.794 | 0.835 | 0.800 | 0.852 |
| +DIR & A$\beta$ value | 0.788 | 0.820 | 0.821 | 0.791 | 0.831 | 0.799 | 0.849 |
| +DIR & A$\beta$ status & APOE4 | 0.796 | 0.822 | 0.824 | 0.795 | 0.834 | 0.802 | 0.851 |
| +DIR & A$\beta$ value & APOE4 | 0.793 | 0.819 | 0.821 | 0.791 | 0.831 | 0.799 | 0.849 |

Fig. S5 shows time-dependent AUC measures of the AIBL MCI subjects at year 2 and year 3 obtained by prediction models built on the deep learning imaging features, shape features, and a combination of shape and texture features (follow-up information at 12 months is not available). The time-dependent AUC measures indicated that the deep learning features had better prediction performance than the conventional shape and texture imaging features. Fig. S6 shows Kaplan-Meier plots of MCI subgroups with different progression risks in terms of progression to AD estimated by the deep learning prediction model on the AIBL cohort. Statistically significant group difference was observed in terms of their progression time to AD dementia (Log-rank test, p<0.009).

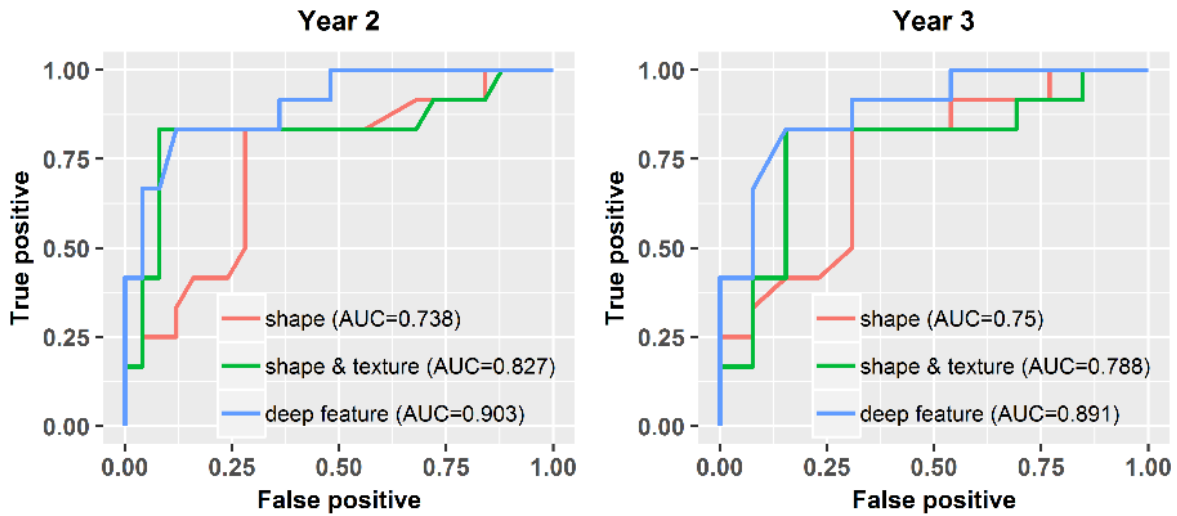

Fig. S5. Time-dependent ROC curves of prognostic models built upon different imaging features at follow-up durations from year 2 to year 3 on the AIBL cohort (clinical outcomes at year 1 is not available in AIBL cohort). The prediction model built on the deep learning features was significantly better than the prediction model built on the shape features in terms of their time dependent AUC values at both year 2 and year 3 with p values of 0.011 and 0.055 respectively. The difference in AUC values was not statistically significant between prediction models built on the deep learning features and the shape and texture features with p values of 0.122 and 0.08 at year 2 and year 3, respectively.

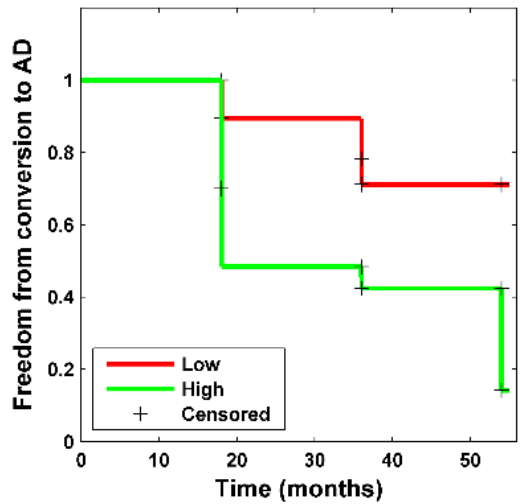

Fig. S6. Kaplan-Meier plot of MCI subgroups with different progression risks in terms of progression to AD dementia estimated by the deep learning prediction model on the AIBL cohort (Low: smaller than the median, High: equal or larger than the median). The subgroups were significantly different in their conversion timing to AD (Log-rank test, p<0.009).

As summarized in Table S1 and Table S2, significant differences were observed between the training data (the ADNI-1 cohort) and the testing data (the ADNI-GO&2 and the AIBL cohorts) in their age, MMSE, and gender ratios. The differences between MCI subjects the ADNI-1 and the ADNI-GO&2 cohorts are probably caused by the inclusion of an early MCI (EMCI) category in the ADNI-GO&2 cohorts. We evaluated the deep learning classification performance based on subsets of the ADNI-GO&2 and AIBL cohorts with matched baseline characteristics with the ADNI-1 cohort. Particularly, younger NC subjects of the ADNI-GO&2 and the AIBL cohorts and AIBL NC and AD subjects had smaller MMSE were excluded in this evaluation. The resulting demographic information of the subsets of the ADNI-GO&2 and the AIBL cohorts is summarized in Table S8.

As shown in Fig. S7, the deep learning classifier had better performance than the RF classifiers built on the hippocampal shape features and a combination of shape and texture features on both the ADNI-GO&2 and AIBL cohorts. However, the overall classification accuracy and AUC measures were smaller than those estimated based on all available NC and AD subjects of the ADNI-GO&2 and AIBL cohorts.

Table S8. Demographic information of the NC and AD subjects in subsets of the ADNI-GO&2 and the AIBL cohorts with matched age, gender, and MMSE measures with the ADNI-1 cohort.

| Cohorts | | NC | AD |
|---|---|---|---|
| | \multicolumn{3}{c}{NC and AD subjects} | |
| ADNI-1 | Number of subjects | 228 | 192 |
| | Gender (M/F) | 118/110 | 101/91 |
| | Age | 75.97±5.02 | 75.34±7.45 |
| | MMSE | 29.11±1.00 | 23.31±2.04 |
| ADNI-GO&2 | Number of subjects | 225 | 158 |
| | Gender (M/F) | 106/119 | 91/67 |
| | Age | 75.27±4.90 | 74.85±8.09 |
| | MMSE | 29.00±1.31 | 23.1±2.07 |
| AIBL | Number of subjects | 178 | 55 |
| | Gender (M/F) | 83/95 | 22/33 |
| | Age | 75.25±4.58 | 73.31±7.92 |
| | MMSE | 29.07±0.82 | 22.49±3.21 |

NC: cognitively normal control; AD: Alzheimer's disease

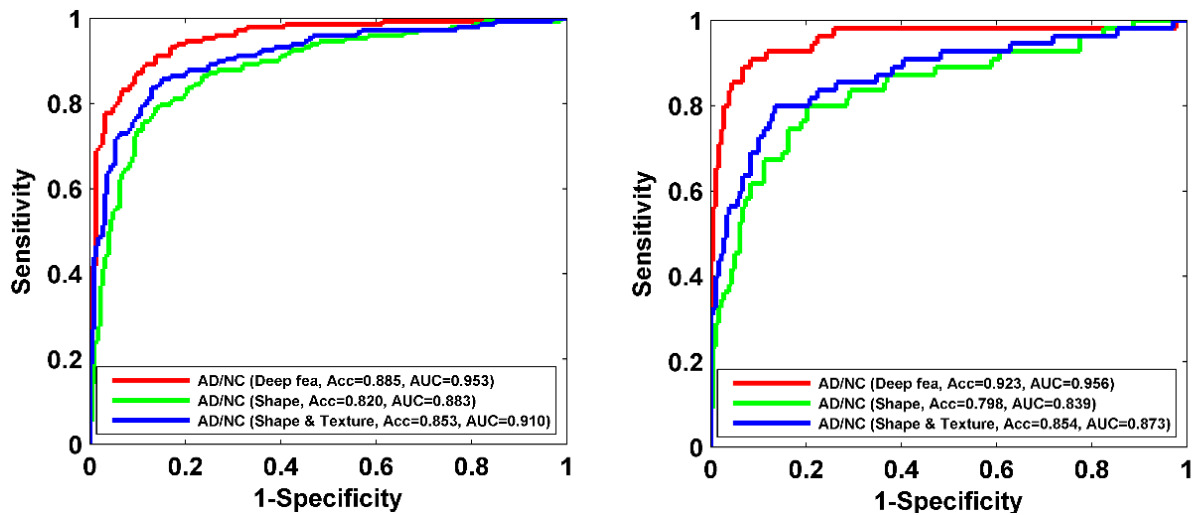

Fig. S7. ROC curves and classification accuracy (Acc) rates obtained by different binary classifiers for distinguishing AD patients from NC subjects, including the deep learning classifier, and RF classifiers built upon the hippocampal shape and texture features, estimated based on subsets of the ADNI-GO&2 (left) and AIBL (right) cohorts with matched age, gender, and MMSE measures.

We further evaluated the deep learning prognostic model based on a subset of the ADNI-GO&2 MCI subjects with matched baseline characteristics with the ADNI-1 MCI subjects. Particularly, younger MCI subjects with larger MMSE of the ADNI-GO&2 cohorts were excluded in this evaluation. The resulting demographic information of the MCI subset of the ADNI-GO&2 cohorts is summarized in Table S9.

As summarized in Table S10, the prediction model built on the deep learning imaging features had better performance than those build on the conventional shape features and a combination of shape and texture features, and the prediction model built on a combination of deep learning features and clinical measures had better performance than the prediction model built on the clinical measures alone. However, the C-index values were smaller than those estimated based on all available MCI subjects of the ADNI-GO&2 cohorts.

In summary, the results shown in Fig. S7 and Table S10 indicated that prediction performance could not be improved by simply removing testing data points to obtain a testing data set matched with the training data set in terms of simple characteristics, such as age and MMSE, indicating that more sophisticated data harmonization techniques are needed and merit further investigation.

Table S9 Demographic and follow-up information of the MCI subset of the ADNI-GO&2 cohorts with matched age, gender, and MMSE measures with the ADNI-1 MCI subjects.

| | MCI subjects with follow-up | | | | | | |
|---|---|---|---|---|---|---|---|
| Cohorts | Number of subjects | Gender (M/F) | Age (years) | MMSE | Follow-up months (quartiles) | Progressors | Months to progression (quartiles) |
| ADNI-1 | 383 | 246/137 | 74.73±7.32 | 27.01±1.77 | 6-132 ([12, 24, 48]) | 208 | 6-126 ([12, 24, 36]) |
| ADNI-GO&2 | 211 | 125/86 | 73.87±6.49 | 26.72±1.34 | 6-78 ([24, 36, 48]) | 107 | 6-60 ([12, 24, 36]) |

MCI: mild cognitive impairment; Converter: MCI subjects who progressed to AD dementia before the last follow-up visit.

Table S10. Prediction performance of prognostic models built upon different types of features, estimated based the MCI subset of the ADNI-GO&2 cohorts with matched age, gender, and MMSE measures with the ADNI-1 MCI subjects.

| C-index / Cohort | Hippocampal imaging features | | | Clinical measures | Clinical measures with deep learning imaging features |
|---|---|---|---|---|---|
| | Shape | Shape & Texture | Deep learning | | |
| ADNI-GO&2 | 0.646 | 0.673 | **0.731** | 0.833 | **0.842** |

The clinical measures included APOE4 status, all cognitive measures, and demographic data.

We have found that 13 (ADNI 1), 32 (ADNI GO&2), and 3 (AIBL) MCI subjects at baseline have reverted to cognitively normal state at their last visit. Therefore, we also evaluated the prognosis models by excluding these subjects. Particularly, we trained and evaluated the prognosis models with these subjects excluded using the same training and evaluation procedure as described in the manuscript. The prognosis performance is summarized in Table S11. The C-index values are similar to those reported in the manuscript. The deep learning based features are more discriminative than the conventional shape and texture features, and could provide complimentary information to clinical measures for prediction. It is noteworthy that the inclusion of MCI subjects with reversion did not violate the assumption of the prognosis model, as they are likely to have lower risk of progression to AD compared to subjects with stable MCI.

Table S11. Prediction performance of prognostic models built upon different types of features after excluding MCI subjects who converted to cognitively normal state at the last visit.

| C-index | Hippocampal imaging features | | | Clinical measures | Clinical measures with deep imaging features |
|---|---|---|---|---|---|
| Cohort | Shape | Texture | Deep learning | | |
| ADNI GO&2 | 0.657 | 0.686 | **0.746** | 0.837 | **0.852** |
| AIBL | 0.648 | 0.750 | **0.822** | | |

### *When volume measures of hippocampus are used for prognosis*

Other than shape based features used in this study, we have also built a prognostic model based on only volume measures of left and right hippocampus using Cox regression for predicting MCI subjects' progression to AD dementia. The volume based model predicted the ADNI-GO&2 MCI subjects' progression to AD dementia with a C-index of 0.651 (A$\beta$-positive subjects' progression to AD with a C-index of 0.631), and the AIBL MCI subjects' progression to AD dementia with a C-index of 0.691, comparable to prediction model built up shape features ($p$=0.239 and 0.344 [7] on the ADNI-GO&2 and the AIBL cohorts respectively) and worse than prediction models built on texture imaging features ($p$=0.018 and 0.104 on the ADNI-GO&2 and the AIBL cohorts respectively) and deep learning imaging features ($p$=0.0005 and 0.124 on the ADNI-GO&2 and the AIBL cohorts respectively). These results also demonstrated that advanced imaging features were more informative for the prognosis.

### *Concordance index (C-index) used for evaluating the prognostic performance*

The C-index is a generalization of the area under the ROC curve (AUC) for validating the predictive ability of a time-to-event analysis model. The C-index was estimated as $C_i = \frac{1}{|P|}\sum_{(i,j)\in P} I(h_r(x_i) < h_r(x_j))$, where $P$ is the set of all orderable pairs of subjects, $|P|$ is the number of pairs in $P$, $h_r(x)$ is a predicted risk of a subject with features of $x$, and $I$ is an indicator function of whether a condition is satisfied or not. Larger C-index indicates better prediction. In the present study, an orderable pair of MCI subjects must have at least one MCI converter. Two MCI subjects who did not progress to AD dementia are not orderable since we do not know which one will progress to AD dementia earlier than the other one. For the MCI subjects of the ADNI-GO&2 cohorts, we have 33739 orderable pairs. The difference of 0.02 between C-index values of two prediction models indicates that the model with a larger C-index value correctly predicted ~675 pairs more than the model with a smaller C-index value.

### *Implementation and feasibility of transferring to other settings*

The whole pipeline of the proposed prognostic model is automatic. Our method is not sensitive to hippocampus segmentation, as only a bounding box containing hippocampus is defined based on the segmentation result. The deep learning feature extraction and prognosis is efficient on both GPU and CPU (within 1 s) once the trained model is obtained. Particularly, for each subject, it takes ~20 mins on a standard PC to segment the hippocampus, it takes ~0.011 s on a GPU or ~0.463 s on a CPU to compute deep learning features, and it takes ~0.1ms to obtain a prognosis result on a CPU.

Source codes and scripts of the hippocampus segmentation method used in the present study are available at www.nitrc.org/frs/?group_id=1242. Comparison results with alternative hippocampus segmentation methods are available at https://doi.org/10.3389/fninf.2018.00069.

The deep learning codes can be used across platforms, including cloud computing, once they are containerized using docker (www.docker.com).